\documentclass{article}
\ifdefined\pdfminorversion
\fi

\usepackage{iclr2027_conference,times}

\usepackage{amsmath,amsfonts,bm}

\def\eqref#1{equation~\ref{#1}}

\def\1{\bm{1}}

\DeclareMathAlphabet{\mathsfit}{\encodingdefault}{\sfdefault}{m}{sl}
\SetMathAlphabet{\mathsfit}{bold}{\encodingdefault}{\sfdefault}{bx}{n}

\usepackage{hyperref}
\hypersetup{hidelinks}
\usepackage{xurl}
\usepackage{etoolbox}
\usepackage{array}
\usepackage{booktabs}
\usepackage{xcolor}
\usepackage{colortbl}
\usepackage{pifont}
\usepackage{graphicx}
\usepackage{placeins}
\usepackage{needspace}

\AtBeginEnvironment{thebibliography}{\raggedright}

\newcommand{\cmark}{\textcolor{green!70!black}{\ding{51}}}
\newcommand{\xmark}{\textcolor{red!70!black}{\ding{55}}}
\newcommand{\method}{\textsc{CoD}}
\newcommand{\TODO}[1]{}
\definecolor{sodteal}{HTML}{16877A}
\definecolor{sodteallight}{HTML}{EDF8F5}

\title{Chain on Demand:\\
Deferred Evidence Organization\\
for Long-Term Conversational Memory}

\author{
Di Lu\textsuperscript{1} \qquad
Zechen Li\textsuperscript{1} \qquad
Hao Xue\textsuperscript{1,2,3}
\\[1ex]
\textsuperscript{1}University of New South Wales, Sydney\\
\textsuperscript{2}Hong Kong University of Science and Technology (Guangzhou)\\
\textsuperscript{3}Hong Kong University of Science and Technology
\\[1ex]
\texttt{di.lu1@student.unsw.edu.au}\\
\texttt{zechen.li@unsw.edu.au}\\
\texttt{haoxue@hkust-gz.edu.cn}
}

\iclrfinalcopy

\begin{document}

\maketitle

\begin{abstract}
Long-term conversational memory faces a timing mismatch: write-time summaries
and graphs organize history before a future query is known, while independent
retrieval can allocate a limited evidence budget to redundant or fragmented
turns. We introduce Chain on Demand (\method{}), a query-time evidence
organization framework over an unchanged turn-level memory. Given a broad
candidate pool, \method{} constructs a temporary evidence chain by allocating
a fixed anchor budget according to query relevance, lexical diversity relative
to selected anchors, and newly covered query terms, followed by deterministic
restoration of each selected anchor's same-session neighborhood. The resulting
chain is used only for the current query, while the underlying conversational
memory remains unchanged. In a paired comparison over all 1,986 LoCoMo
questions, changing only the allocation rule improves accuracy from 74.5\%
to 76.3\% (+1.86 percentage points; conversation-cluster 95\% CI
[0.59, 3.15]). \method{} also achieves 82.2\% accuracy on the 1,540
non-adversarial questions. On LongMemEval-S, one fixed
configuration that indexes both user and assistant turns obtains 86.6\% across
all 500 questions. These results show that long-term memory can preserve the
original conversation while deferring evidence organization until the
information need is known. Our code is available at https://github.com/Renkiroo/chainondemand
\end{abstract}

\section{Introduction}
\label{sec:introduction}

Conversational agents need to retain information across interactions that span
days, weeks, or months. A later interaction may depend on a fact mentioned
much earlier, a preference established in another session, or several pieces
of information spread across the conversation. As history grows, the
information needed for a later question becomes harder to recover from the
full interaction record. Since the complete history rarely fits within the
effective context available to the answer model, long-term memory systems
must determine which parts of the history to present for the current query.

Long-term memory involves two separable decisions: what representation
persists across interactions, and what evidence is presented for a particular
query. Many systems construct summaries, segments, links, graphs, or
hierarchical memory units during or shortly after interaction
\citep{zhong2023memorybankenhancinglargelanguage,ICLR2025_e56f394b,
NEURIPS2025_19909c36,rasmussen2025zeptemporalknowledgegraph,
ICLR2026_a05b7265,xu-etal-2026-structmem,ICLR2026_93e444b4}. These
representations can make long histories easier to search and manage. However,
they commit to an organization before the later query reveals which details
matter. Retaining the original turns avoids rewriting the conversational
record, but does not determine how a limited query-time evidence budget should
be allocated. Persistent storage and query-time evidence construction
therefore serve different objectives.

Independent query-time retrieval addresses this timing problem only partly.
Standard top-$K$ retrieval scores candidates independently
\citep{gao2024retrievalaugmentedgenerationlargelanguage}. Under a fixed
budget, several similar turns may occupy the available slots while a less
similar turn supplies a missing relation, temporal detail, or intermediate
step. The retrieved turns can therefore be individually relevant but
collectively insufficient. Query-conditioned compression and set-wise
retrieval account for query requirements more explicitly
\citep{ICLR2024_bda88ed2,lee-etal-2025-shifting}, but conversational evidence
adds another dependency: a selected turn may require nearby dialogue and
temporal order to be interpreted. Simply retrieving more text does not
necessarily resolve this issue, as models can fail to use relevant information
embedded in long inputs~\citep{liu-etal-2024-lost}. Recent conversational
memory systems recover source context through temporal--semantic expansion,
archive-level context recovery, or reranking and tool use
\citep{wang2026cuememcueguidedcontextreconstruction,
li2026agentopenschatapp,sen2026chronostemporalawareconversationalagents}.
These approaches establish the value of source-context recovery, but leave
open a central question: given a fixed candidate pool and anchor budget, how
should conversation locations be selected jointly before their local context
is reconstructed?

We introduce Chain on Demand (\method{}), a query-time evidence organization
framework that constructs a temporary evidence chain over an unchanged turn
record. Given a query, \method{} first retrieves a broad candidate pool, then
allocates a fixed anchor budget using a set-aware relevance--diversity--coverage
rule, and finally restores the original same-session neighborhood around the
selected locations. The selected anchors therefore specify where evidence
should be organized rather than constituting the final evidence context
themselves. The resulting chain is constructed only for the current
information need and discarded after answering, while the underlying
conversational record remains unchanged. This separates persistent memory
from query-specific evidence construction while making the allocation
decision explicit and directly testable. An adaptive acquisition variant
further allows unresolved evidence gaps to trigger additional query-time
selection without modifying the persistent memory.

Table~\ref{tab:design_comparison} positions \method{} relative to
representative query-time evidence construction designs. The comparison
highlights three properties that jointly define our framework: preserving the
original conversational record, using a fixed query procedure with explicit
set-aware allocation, and restoring source context after selection.

\begin{table}[t]
\centering
\caption{Query-time evidence construction in representative systems. ``No
generated semantic memory'' excludes query-independent summaries, cues, and
events; sparse or dense indexes do not count. ``Fixed query procedure'' excludes
multi-round agent search. ``Set-aware allocation'' evaluates each candidate
against the evolving selected set. ``Source-context restoration'' adds original
neighboring turns after selection. The columns describe mechanisms rather than
an ordinal ranking. $\dagger$ denotes document-level RAG.}
\scriptsize
\renewcommand{\arraystretch}{1.18}
\setlength{\tabcolsep}{1.5pt}
\begin{tabular*}{\linewidth}{@{\extracolsep{\fill}}lcccc@{}}
\toprule
\textbf{Design}
& \shortstack{No generated\\semantic\\memory}
& \shortstack{Fixed query\\procedure}
& \shortstack{Set-aware\\allocation}
& \shortstack{Source-context\\restoration} \\
\midrule
LoCoMo dialog RAG~\citep{maharana-etal-2024-evaluating}
& \cmark & \cmark & \xmark & \xmark \\
\addlinespace[1pt]
SeCom~\citep{ICLR2025_e56f394b}
& \xmark & \cmark & \xmark & \xmark \\
\addlinespace[1pt]
SetR$^{\dagger}$~\citep{lee-etal-2025-shifting}
& -- & \cmark & \cmark & \xmark \\
CueMem~\citep{wang2026cuememcueguidedcontextreconstruction}
& \xmark & \cmark & \xmark & \cmark \\
ReFind~\citep{li2026agentopenschatapp}
& \cmark & \xmark & \xmark & \cmark \\
Chronos~\citep{sen2026chronostemporalawareconversationalagents}
& \xmark & \xmark & \xmark & \cmark \\
\addlinespace[2pt]
\rowcolor{sodteallight}
\textbf{\method{} (ours)}
& \cmark & \cmark & \cmark & \cmark \\
\bottomrule
\end{tabular*}
\vspace{-1em}
\label{tab:design_comparison}
\end{table}

Our central evidence is a paired LoCoMo comparison that changes only the
allocation rule. Under the same memory, candidate pool, anchor budget,
temporal reconstruction, answer model, and judge, \method{} raises accuracy
from 74.5\% to 76.3\%, a 1.86-point gain with a conversation-cluster 95\%
confidence interval of [0.59, 3.15]. Development diagnostics further isolate
the contributions of evidence volume, temporal reconstruction, and
set-aware allocation. On LongMemEval-S, a fixed configuration that indexes
both user and assistant turns obtains 433/500 (86.6\%) across all questions,
supporting applicability on a second long-term memory benchmark.

Our contributions are threefold:
\begin{itemize}
\item We formulate long-term conversational memory around deferred evidence
organization: the persistent conversational record is preserved, while
query-specific evidence is constructed only after the information need is
known.

\item We introduce \method{}, a query-time evidence organization framework
that constructs a temporary evidence chain through broad candidate retrieval,
set-aware fixed-budget anchor allocation, and same-session context
reconstruction. An adaptive acquisition variant further allows unresolved
evidence gaps to trigger additional selection.

\item We isolate the effect of set-aware allocation in a paired comparison over
all 1,986 LoCoMo questions: changing only the allocation rule improves accuracy
from 74.5\% to 76.3\% (+1.86 points; conversation-cluster 95\% CI
[0.59, 3.15]). Development analyses characterize reconstruction and allocation,
and LongMemEval-S evaluates the complete pipeline on a second benchmark.
\end{itemize}
\section{Related Work}
\label{sec:related-work}

\paragraph{Persistent Memory Construction.}
Long-term conversational memory systems differ in what they preserve across
interactions. MemoryBank summarizes events and user traits
\citep{zhong2023memorybankenhancinglargelanguage}; SeCom builds topical segments
and compressed retrieval units~\citep{ICLR2025_e56f394b}; and A-MEM augments
stored interactions with attributes, links, and evolving notes
\citep{NEURIPS2025_19909c36}. Other systems use temporal graphs, structured
summaries, or multiple levels of granularity
\citep{rasmussen2025zeptemporalknowledgegraph,xu-etal-2026-structmem,
ICLR2026_a05b7265,kang-etal-2025-memory,ICLR2026_93e444b4}. These designs
primarily determine what information persists and how it is indexed. Even when
they route or filter memory according to a query, the underlying textual units
or relations are constructed before that query is known. \method{} instead
keeps the original turn record unchanged and defers the allocation of
query-specific evidence until the information need is available.

\paragraph{Query-Time Evidence Allocation and Reconstruction.}
Standard retrieval-augmented generation independently ranks candidates
\citep{gao2024retrievalaugmentedgenerationlargelanguage}. MMR introduced
relevance--diversity reranking~\citep{Carbonell_1998}; RECOMP performs
query-conditioned compression~\citep{ICLR2024_bda88ed2}; and SetR explicitly
selects passages as a set~\citep{lee-etal-2025-shifting}. Multi-stage and
agentic systems make retrieval or memory operations adaptive through
deliberative search, learned policies, or program evolution
\citep{cao-etal-2026-higmem,you2026dmemdualprocessmemoryllm,packer2023memgpt,
yan2026memory,yu2026agentic,liu2026memproagenticmemorysystems,
li-etal-2026-zara}.

In conversational memory, the closest systems already restore source context.
CueMem independently retrieves generated cues, maps them to source-turn
anchors, and expands over a precomputed temporal--semantic graph
\citep{wang2026cuememcueguidedcontextreconstruction}. ReFind combines an
unmodified chat archive with multi-round agent-controlled lexical search and
local context expansion~\citep{li2026agentopenschatapp}. Chronos combines raw
turn and extracted event calendars with reranking, same-session expansion, and
an iterative tool loop
\citep{sen2026chronostemporalawareconversationalagents}. These lines of work
provide query conditioning, raw-record access, set selection, and local
expansion in different combinations. \method{} combines these components into
a query-time pipeline in which selected units are conversation locations for
subsequent source reconstruction rather than final passages. Its allocator is
fixed and non-agentic, allowing its contribution to be measured while holding
the candidate pool, anchor budget, reconstruction, answer model, and judge
fixed.

\section{Methodology}
\label{sec:method}

Given a query $q$ and an ordered memory of original conversation turns,
Chain on Demand (\method{}) constructs query-specific evidence at query time.
The pipeline separates three decisions: which turns are available, which
conversation locations receive a fixed anchor budget, and which original
context is restored around those locations. Its core allocation mechanism
explicitly accounts for the evolving selected set: a greedy rule re-scores
candidates as anchors are added. Candidate access defines the feasible
locations, while deterministic reconstruction maps selected locations back to
source context without making another set-dependent allocation decision.

\begin{figure}[t]
    \centering
    \includegraphics[width=\linewidth]{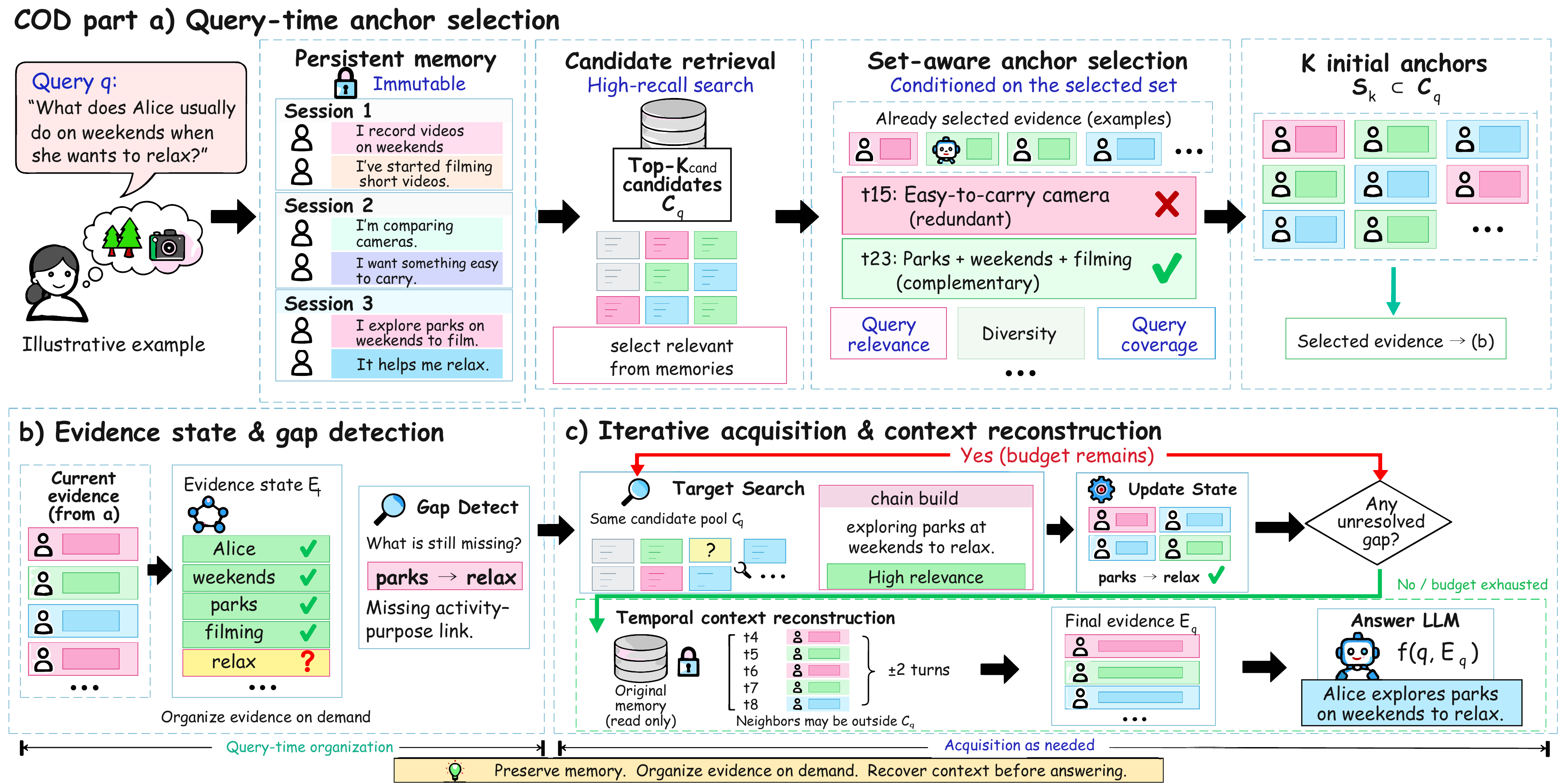}
    \caption{\textbf{Chain on Demand.} (a) Query-time set-aware allocation
    selects anchor locations from a fixed candidate pool, followed by
    deterministic same-session reconstruction. (b--c) An adaptive gap-directed
    acquisition variant iteratively updates a temporary evidence state and
    acquires additional anchors within the same candidate pool. The end-to-end
    results use the fixed-budget pipeline; the adaptive variant is evaluated
    separately with selection-level diagnostics.}
    \label{fig:cod-overview}
\end{figure}

Let $\mathcal{M}=(m_1,\ldots,m_N)$ denote the persistent memory, where each turn
$m_i=(x_i,r_i,s_i,\tau_i)$ contains text $x_i$, speaker role $r_i$, session
identifier $s_i$, and session-local position $\tau_i$. \method{} never rewrites
$x_i$, and its temporary evidence is not written back to memory.
Figure~\ref{fig:cod-overview} distinguishes the fixed-budget core pipeline from
the adaptive gap-directed acquisition variant.

\subsection{Candidate Access}
\label{sec:candidate-retrieval}

Candidate retrieval determines which turns are available to the allocator; it
does not determine the final evidence. Each stored turn has a dense embedding,
and the query retrieves
\begin{equation}
\mathcal{C}_q=
\operatorname{TopK}_{m_i\in\mathcal{M}}
\left(
R(m_i,q);K_{\mathrm{cand}}
\right),
\qquad
R(m_i,q)=s_{\mathrm{dense}}(m_i,q).
\label{eq:candidate-retrieval}
\end{equation}
The resulting pool $\mathcal{C}_q$ defines the fixed search space for anchor
allocation. Candidate access therefore determines evidence availability,
while the subsequent allocator determines how the anchor budget is
distributed.

\subsection{Fixed-Budget Set-Aware Anchor Allocation}
\label{sec:anchor-selection}

Independent top-$K$ retrieval assigns each turn a fixed score. Once one turn is
selected, those scores do not reflect what the selected set already contains.
\method{} instead recomputes a candidate's score after every selection.
Starting from $S_0=\emptyset$, it applies the greedy update
\begin{equation}
\begin{aligned}
u_t(m;q)
&=
R(m,q)
+\lambda D(m,S_t)
+\beta C(m,q,S_t),\\
m_{t+1}^{*}
&=
\arg\max_{m\in\mathcal{C}_q\setminus S_t}u_t(m;q),
\qquad
S_{t+1}=S_t\cup\{m_{t+1}^{*}\}.
\end{aligned}
\label{eq:anchor-selection}
\end{equation}
Here, $R$ is dense query relevance. The lexical diversity term
\begin{equation}
D(m,S_t)=
1-\max_{s\in S_t}
\operatorname{Jaccard}
\bigl(
\operatorname{tok}(m),\operatorname{tok}(s)
\bigr)
\label{eq:diversity}
\end{equation}
penalizes overlap with selected anchors, with $D(m,\emptyset)=1$. The coverage
term $C(m,q,S_t)$ is the proportion of currently uncovered query terms matched
by $m$; it is zero when no query terms remain uncovered.

The update repeats until $|S_t|=K_{\mathrm{anchor}}$. Because $D$ and $C$
depend on $S_t$, the score of an unselected candidate can change after every
choice. This is the defining set-aware operation in \method{}. The resulting
turns are \emph{anchors}: they identify conversation locations from which the
query-specific evidence chain is subsequently constructed rather than serving
directly as the final evidence context.

\subsection{Deterministic Same-Session Reconstruction}
\label{sec:temporal-reconstruction}

An anchor can identify the relevant location while omitting dialogue needed to
interpret it. For each selected anchor $m_i$, \method{} restores a fixed
neighborhood from the original memory:
\begin{equation}
\mathcal{N}_w(m_i)
=
\{m_j\in\mathcal{M}\mid s_j=s_i,\ |\tau_j-\tau_i|\leq w\}.
\label{eq:temporal-neighborhood}
\end{equation}
The final evidence $\mathcal{E}_q$ is obtained by taking the union of the
neighborhoods of all selected anchors, removing duplicate turns, and
preserving their original chronological order. The neighborhoods are read
from $\mathcal{M}$ rather than $\mathcal{C}_q$, so a restored neighbor need not
belong to the candidate pool.

Conditioned on the anchors and window width, reconstruction performs only
same-session indexing, deduplication, and chronological ordering. It adds no
semantic retrieval, generated memory, or further allocation decision. The
anchors therefore define the locations from which the temporary query-specific
evidence chain is constructed. The answer model receives $q$ and
$\mathcal{E}_q$, while the persistent memory remains unchanged.

\paragraph{Adaptive Gap-Directed Acquisition.}
\label{sec:gap-directed-method}
Figure~\ref{fig:cod-overview} also illustrates an adaptive variant that updates
a temporary evidence state while reallocating anchors within the same fixed
candidate pool. The procedure can acquire additional evidence when the current
state leaves unresolved gaps. We evaluate this variant separately with
selection-level diagnostics; it is not included in the end-to-end results of
the fixed-budget pipeline. Appendix~\ref{sec:agentic-extension} reports its
configuration and diagnostic results.
\section{Experiments}
\label{sec:experiments}

We evaluate \method{} on LoCoMo~\citep{maharana-etal-2024-evaluating} and
LongMemEval-S~\citep{wu2025longmemevalbenchmarkingchatassistants}. Our main
comparison isolates the effect of set-aware allocation by fixing candidate
access, anchor count, reconstruction, answer model, and judge. We further
evaluate performance across question types, analyze the roles of evidence
volume and temporal reconstruction on a 100-question LoCoMo development
subset, and measure the query-time cost of the evidence-construction
pipeline.

\subsection{Experimental Setup}
\label{sec:exp_setup}

For LoCoMo, we report results on both the 1,540 non-adversarial questions and
the full set of 1,986 questions. For LongMemEval-S, we evaluate all 500
questions. GPT-4o-mini serves as both the answer model and evaluator in our
runs; answer generation uses temperature 0 and at most 512 tokens.

We retain each original turn with its speaker, timestamp, session, and dense
embedding. Using \texttt{text-embedding-3-small} and Qdrant, \method{} retrieves
200 candidates, allocates 60 anchors with $\lambda=0.06$ and $\beta=0.10$, and
restores two turns on each side within the same session. LongMemEval-S indexes
both user and assistant turns.

The primary comparison changes only the allocator; memory, questions,
candidate pool, anchor budget, reconstruction, answer model, and judge remain
fixed. Mechanism tests use a fixed 100-question LoCoMo development subset that
is contained in the 1,986-question set, so these results are diagnostics
rather than independent held-out estimates. Published values are reproduced
from MemPro's comparison table~\citep{liu2026memproagenticmemorysystems}, which
combines reproduced held-out runs with values taken from prior papers. MemPro
reserves 154/1,540 LoCoMo and 50/500 LongMemEval questions for framework
evolution and evaluates on the remaining 1,386 and 450; our rows report all
1,540 and 500. These values therefore provide context rather than matched
comparisons. Appendix~\ref{sec:additional-details} gives the full protocol.

\subsection{Main Results}
\label{sec:main_results}

The matched comparison changes only the allocation rule while keeping the
memory, candidate pool, 60-anchor budget, reconstruction, answer model, and
judge fixed. On all 1,986 LoCoMo questions, \method{} improves accuracy from
74.5\% with relevance-only allocation to 76.3\% (+1.86 points), with a
conversation-cluster 95\% CI of [0.59, 3.15]. The paired question-level
comparison gives 135 wins and 98 losses for \method{} (Appendix
\ref{sec:significance}). This comparison directly measures the effect of
set-aware allocation under the same evidence-construction pipeline.

\begin{table}[t]
\centering
\caption{Cross-protocol LoCoMo context with GPT-4o-mini (\%). \method{} is
evaluated on all 1,540 non-adversarial questions. Multi, Open, Single, and
Temp denote multi-hop, open-domain, single-hop, and temporal reasoning.}
\small
\renewcommand{\arraystretch}{1.05}
\setlength{\tabcolsep}{6pt}

\begin{tabular}{lccccc}
\toprule
\textbf{Method}
& \textbf{Overall} $\uparrow$
& \textbf{Multi}
& \textbf{Open}
& \textbf{Single}
& \textbf{Temp} \\
\midrule

\textcolor{gray}{Full Text}
& 73.83 & 68.79 & 56.25
& 86.56 & 50.16 \\

\textcolor{gray}{RAG}
& 63.64 & 55.32 & 47.92 & 70.99 & 56.39 \\

\midrule

\textcolor{violet}{LangMem}
& 55.45 & 52.10 & 41.65
& 62.80 & 43.25 \\

\textcolor{violet}{Mem0}
& 36.49 & 30.85 & 34.38
& 38.41 & 37.07 \\

\textcolor{violet}{A-MEM}
& 64.16 & 56.03 & 31.25
& 72.06 & 60.44 \\

\textcolor{violet}{MemoryOS}
& 58.25 & 56.74 & 45.83
& 67.06 & 40.19 \\

\midrule

\textcolor{olive}{LightMem}
& 70.26 & 62.06 & 42.71
& 74.67 & 74.14 \\

\textcolor{olive}{SimpleMem}
& 72.08 & 64.50 & 44.90
& 76.80 & 74.50 \\

\midrule

\rowcolor{green!8}
\textbf{\method{} (ours)}
& 82.20
& 71.30
& 58.30
& 88.70
& 81.90 \\

\bottomrule
\end{tabular}

\label{tab:locomo}
\vspace{-0.5em}
\end{table}

On the non-adversarial LoCoMo split, \method{} achieves 82.2\% accuracy
(1,266/1,540), and it achieves 76.3\% (1,516/1,986) on the complete benchmark.
Across question types, the accuracy is 71.3\% for multi-hop, 58.3\% for
open-domain, 88.7\% for single-hop, and 81.9\% for temporal questions.
Table~\ref{tab:locomo} places these results alongside selected published
results. Since those rows are not protocol-matched, the paired comparison
above provides the evidence for the effect of the allocation rule.

\begin{table}[t]
\centering
\caption{Cross-protocol LongMemEval-S context with GPT-4o-mini (\%). The
\method{} row comes from one fixed configuration that
indexes all turns and is applied to all 500 questions. 
Temp, Multi, Know, User, Asst., and Pref. denote
temporal reasoning, multi-session reasoning, knowledge
update, single-user recall, single-assistant recall, and
preference.}
\small
\renewcommand{\arraystretch}{1.05}
\setlength{\tabcolsep}{6pt}

\begin{tabular}{lccccccc}
\toprule
\textbf{Method}
& \textbf{Temp}
& \textbf{Multi}
& \textbf{Know}
& \textbf{User}
& \textbf{Asst.}
& \textbf{Pref.}
& \textbf{Overall} $\uparrow$ \\
\midrule

\textcolor{gray}{Full Text}
& 31.58 & 45.45 & 76.92 & 87.14 & 89.29 & 36.67 & 56.89 \\

\textcolor{gray}{RAG}
& 39.85 & 48.48 & 67.95 & 90.00 & 98.21
& 53.33 & 60.90 \\

\midrule

\textcolor{violet}{LangMem}
& 15.79 & 20.30 & 66.67 & 60.00 & 46.43 & 60.00 & 37.20 \\

\textcolor{violet}{Mem0}
& 40.15 & 46.21 & 70.12 & 81.43 & 41.07 & 60.00 & 53.51 \\

\textcolor{violet}{A-MEM}
& 47.36 & 48.87 & 64.11 & 92.86 & 96.43
& 46.67 & 62.20 \\

\textcolor{violet}{MemoryOS}
& 32.33 & 31.06 & 48.72 & 80.00 & 64.29 & 30.00 & 44.66 \\

\midrule

\textcolor{olive}{LightMem}
& 67.18 & 71.74 & 83.12
& 87.14 & 32.14 & 68.18 & 69.81 \\

\textcolor{olive}{SimpleMem}
& 69.17 & 60.90 & 78.21 & 85.71 & 75.00
& 73.33 & 71.60 \\

\midrule

\rowcolor{green!8}
\textbf{\method{} (ours)}
& 91.70
& 84.20
& 91.00
& 98.60
& 78.60
& 50.00
& 86.60 \\

\bottomrule
\end{tabular}
\label{tab:longmemeval}
\vspace{-1em}
\end{table}

On LongMemEval-S, one fixed \method{} configuration that indexes both user and
assistant turns obtains 86.6\% accuracy (433/500). The corresponding accuracies
are 91.7\% for temporal reasoning, 84.2\% for multi-session reasoning, 91.0\%
for knowledge updates, 98.6\% for single-user recall, 78.6\% for
single-assistant recall, and 50.0\% for preferences. The two benchmarks provide
complementary evidence: the paired LoCoMo comparison isolates the allocator's
effect, while the unified LongMemEval-S run evaluates one fixed
evidence-construction configuration across different question types. Since
LongMemEval-S does not include a matched relevance-only allocator ablation,
these results provide cross-paper context rather than evidence of an allocator
improvement.

On the 56 assistant-memory questions, dual-role indexing raises accuracy from
11/56 (19.6\%) with user-only indexing to 44/56 (78.6\%). This diagnostic
indicates that access to assistant turns matters for these questions in the
evaluated pipeline. It does not test the set-aware allocator and is not used
to construct the unified 500-question result.

In our measured implementation on the same 100 queries, caching token sets and
pairwise overlaps reduces selection time from 7,046 to 115\,ms/query and total
memory-construction time from 7,183 to 262\,ms/query, without changing the
resulting evidence statistics. Dense-60 retrieval takes 21\,ms/query, so the
optimized evidence-construction pipeline adds 241\,ms/query. \method{} uses
13,218 prompt tokens per query versus 4,206 for Dense-60. This comparison
reports the footprint of the full pipeline rather than allocator-only
overhead. Detailed latency and token-usage measurements appear in
Appendices~\ref{sec:efficiency-details} and \ref{sec:api-usage}.

\subsection{Controlled Mechanism Tests}
\label{sec:controlled-analysis}

We next examine three factors on the fixed 100-question LoCoMo development
diagnostic: evidence volume, local reconstruction, and set-aware allocation.
These results are intended to explain the mechanism and are not independent
benchmark estimates.

\begin{table}[t]
\centering
\caption{Evidence construction on the LoCoMo diagnostic (\%, $N=100$).}
\small
\renewcommand{\arraystretch}{1.05}
\setlength{\tabcolsep}{6pt}

\begin{tabular}{lccc}
\toprule
\textbf{Method}
& \textbf{Acc.} $\uparrow$
& \textbf{Turns}
& \textbf{Words} \\
\midrule

Dense-60
& 51.0 & 60.0 & 1,268 \\

Dense-200
& 57.0 & 200.0 & 4,587 \\

Dense + Temporal
& 73.0 & 202.87 & 4,566 \\

\rowcolor{green!8}
\textbf{\method{}}
& \textbf{77.0} & 202.82 & 4,575 \\

\bottomrule
\end{tabular}
\label{tab:evidence_budget}
\end{table}

Increasing independent dense retrieval from 60 to 200 turns raises accuracy
from 51.0\% to 57.0\%. At nearly the same final volume, restoring local
same-session context around relevance-selected anchors reaches 73.0\% (202.87
turns and 4,566 words, compared with 200 turns and 4,587 words). Thus, on this
diagnostic, increasing independent retrieval to a similar evidence volume does
not reproduce the effect of local reconstruction.

Dense plus temporal and \method{} use the same candidate pool, 60-anchor budget,
and reconstruction procedure, and produce nearly identical context sizes:
202.87 versus 202.82 turns and 4,566 versus 4,575 words. Replacing
relevance-only anchors with set-aware allocation increases accuracy from 73.0\%
to 77.0\% at essentially unchanged evidence volume. The resulting comparisons
separate evidence quantity from organization: Dense-60 versus Dense-200 gives
a 6-point increase from broader independent retrieval, Dense-200 versus Dense
plus Temporal gives a 16-point increase from local reconstruction at similar
volume, and Dense plus Temporal versus \method{} gives a 4-point increase from
changing the anchor allocation. These development results are not an additive
decomposition of benchmark-wide performance. Section~\ref{sec:analysis}
further examines diversity, coverage, candidate availability, and alternative
allocation rules.
\section{Mechanism Analysis}
\label{sec:analysis}

Section~\ref{sec:experiments} provides a matched result over all 1,986 LoCoMo
questions showing that set-aware allocation improves answer accuracy. We now
use the fixed 100-question LoCoMo development diagnostic to examine where
query-time evidence construction creates value. We consider local source
reconstruction, the components of set-aware allocation, candidate availability,
and the adaptive acquisition variant. These analyses explain the mechanism;
they are not independent held-out estimates.

\begin{figure*}[t]
    \centering
    \includegraphics[width=0.92\linewidth]{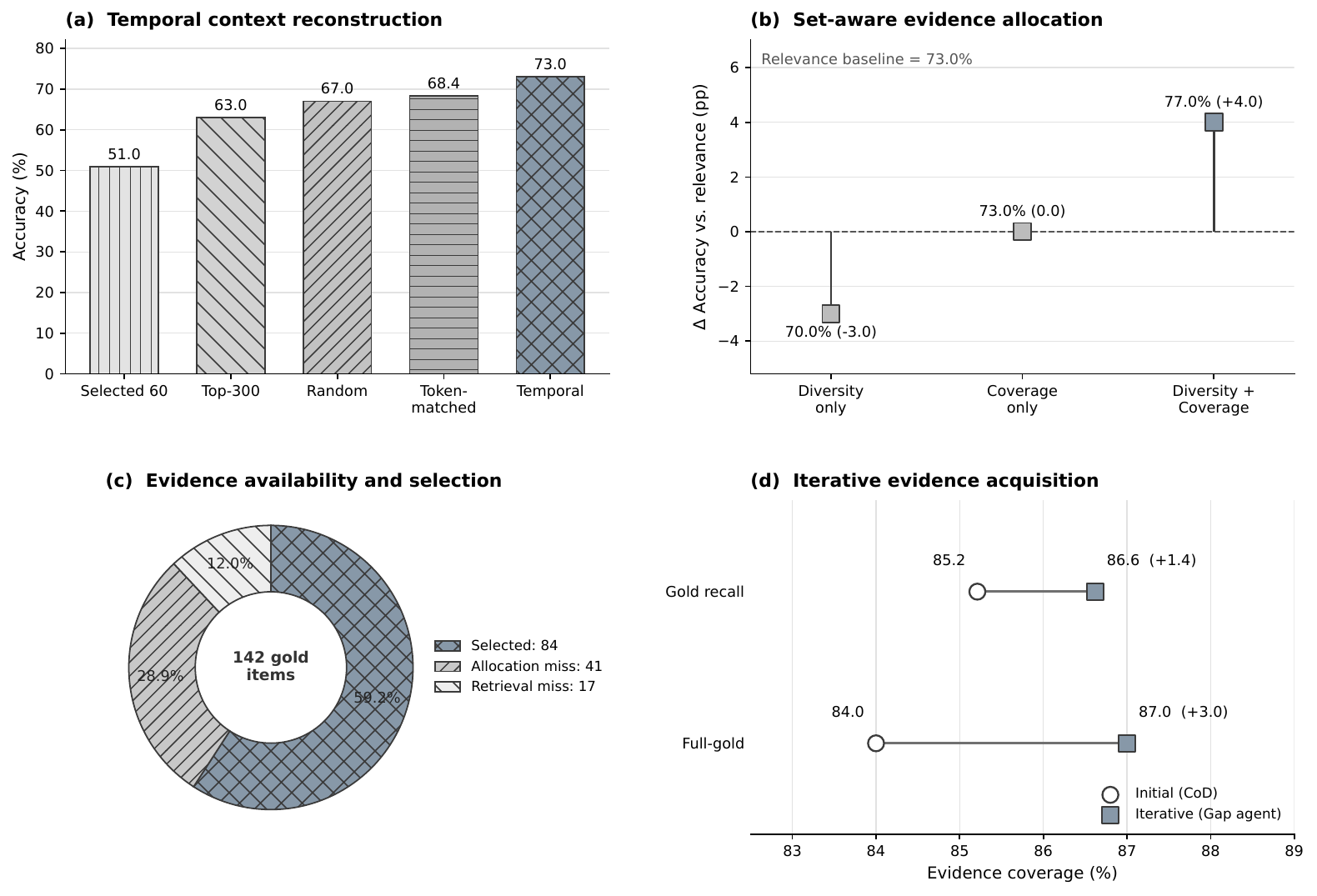}
    \caption{Analysis of query-time evidence construction on the 100-question
    diagnostic subset. (a) Temporal reconstruction improves over independent
    retrieval and random expansion at similar evidence volume. (b) Combining
    diversity and coverage with relevance improves set-aware allocation over
    the relevance-only baseline. (c) Gold evidence is missed both during
    candidate retrieval and anchor allocation. (d) Iterative evidence
    acquisition improves evidence coverage in the selection-only diagnostic.}
    \label{fig:analysis-ablation}
\end{figure*}

\subsection{Local Source Context Helps Beyond Evidence Volume}

Figure~\ref{fig:analysis-ablation}(a) compares independent retrieval,
random expansion, and temporal reconstruction starting from the same 60
relevance-only anchors. Top-300 raw retrieval is a separate
independent-retrieval control. Random expansion samples non-anchor turns from
the full conversation; its token-matched version samples to approximately
match the reconstructed evidence volume and reports the mean over five seeds
(42--46).

Accuracy is 63.0\% for Top-300 raw retrieval, 67.0\% for random expansion,
68.4\% for token-matched random expansion, and 73.0\% for temporal
reconstruction. The 4.6-point gap to the approximately token-matched control
shows that matching evidence volume alone does not reproduce the result.
Instead, restoring the local same-session context around relevance-selected
anchors provides additional value. This comparison supports temporal
reconstruction as a pipeline component but does not test the allocator.

\subsection{Set-Aware Allocation}

Figure~\ref{fig:analysis-ablation}(b) isolates the components of the allocation
rule. With the same candidate pool, anchor budget, reconstruction, answer
model, and judge, relevance-only selection reaches 73.0\%. Adding diversity
alone ($R+D$) reaches 70.0\%, adding coverage alone ($R+C$) reaches 73.0\%,
while the combined $R+D+C$ rule reaches 77.0\%. The diversity and coverage
terms are recomputed against the evolving selected set rather than applied as
fixed per-turn scores.

We further test whether the combined rule needs to control the full anchor
budget. The protected-coverage and protected-$D+C$ variants reach 72.0\% and
73.0\%, respectively, compared with 73.0\% for relevance-only selection.
Neither reproduces the 77.0\% accuracy of \method{}, where relevance,
diversity, and coverage jointly influence all 60 anchor decisions. A
gold-informed allocator reaches 78.0\%, one point above \method{}, but it
directly uses gold evidence identities and is not deployable; we therefore
use it only as a reference.

Together, these controls indicate that the diagnostic improvement is not
reproduced by applying diversity or coverage only after a relevance-ranked
prefix has been fixed. The full set-aware rule allows the three signals to
interact throughout anchor allocation. The corresponding matched comparison
over all 1,986 questions gives a +1.86-point gain with a conversation-cluster
95\% CI of [0.59, 3.15] (Section~\ref{sec:experiments}).

\subsection{Candidate Availability Leaves an Allocation Opportunity}

Figure~\ref{fig:analysis-ablation}(c) tracks the 142 dataset-provided gold
evidence items before reconstruction. The candidate pool contains 125 items
(88.0\%); 84 become anchors, 41 are available but not selected, and 17 are
absent from the pool. Among the 58 items not selected as anchors, 41 (70.7\%)
are therefore already visible to the allocator.

This leaves a substantial allocation opportunity: most gold items missed at
the anchor stage are present in the candidate pool rather than unavailable
to retrieval. At the same time, 17 gold items remain absent from the pool.
These are anchor-stage counts, not final-context misses or answer errors,
since reconstruction can still restore an unselected turn as a neighbor of
another anchor.

\subsection{Adaptive Evidence Acquisition}

Figure~\ref{fig:analysis-ablation}(d) evaluates the adaptive gap-directed
acquisition variant using selection-level evidence coverage. Starting from
the same initial configuration, iterative acquisition improves Gold
Recall@60 from 85.2\% to 86.6\% and Full-Gold Coverage from 84.0\% to
87.0\%. The procedure maintains a temporary evidence state and acquires
additional anchors when unresolved evidence gaps remain, while keeping the
candidate pool fixed.

These results show that additional query-time acquisition can improve
evidence coverage without modifying the persistent memory. They are reported
only as selection-level diagnostics: the end-to-end accuracy results in
Section~\ref{sec:experiments} use the fixed-budget pipeline without the
iterative loop. Appendix~\ref{sec:agentic-extension} gives the full
configuration and diagnostic protocol.
\section{Conclusion}
\label{sec:conclusion}

We introduced Chain on Demand (\method{}), a query-time evidence construction
framework that preserves the original conversational record and organizes
evidence only after the information need is known. Given a broad candidate
pool, \method{} allocates a fixed anchor budget with set-aware greedy scoring
and then reconstructs the original same-session context around the selected
locations, forming a temporary evidence chain without modifying persistent
memory.

On LoCoMo, changing only the allocation rule improves accuracy from 74.5\% to
76.3\% over all 1,986 questions (+1.86 points; conversation-cluster 95\% CI
[0.59, 3.15]). The development analyses further show the complementary roles
of set-aware allocation and source-context reconstruction, while the
LongMemEval-S results demonstrate that the same query-time evidence
construction framework transfers to a second long-term memory benchmark.
Together, these results support deferring evidence organization until the
query is known, rather than committing the organization of conversational
history in advance.

\section{Limitations}
\label{sec:limitations}

Our evaluation has several limitations. First, the 100-question LoCoMo
development subset is contained in the full 1,986-question evaluation, so the
component analyses are development diagnostics rather than independent
held-out estimates. Second, the answer model and judge are fixed to
GPT-4o-mini in the reported runs; other models and evaluators remain
untested. Third, LongMemEval-S does not include a matched relevance-only
allocator ablation, so its results support applicability across benchmarks
rather than an independent estimate of the allocator's effect.

The current configuration also fixes the candidate pool at 200 turns, the
anchor budget at 60, and the reconstruction window at two turns on each side.
We do not systematically study how these choices scale with longer histories
or different evidence budgets. Query-time construction additionally incurs
higher prompt-token and construction costs than Dense-60 in our measurements.
Finally, reconstruction is limited to local same-session context and does not
explicitly model long-range temporal dependencies or cross-session links.

\bibliography{iclr2027_conference}
\bibliographystyle{iclr2027_conference}

\appendix

\section{Additional Experimental Details}
\label{sec:additional-details}

This appendix documents the evaluation scope, core \method{} configuration,
controlled protocols, efficiency, statistical analysis, and additional
diagnostics. The evidence hierarchy follows the main paper: the matched
1,986-question LoCoMo comparison supports the allocator claim; the fixed
100-question subset provides development diagnostics; LongMemEval-S evaluates
the complete pipeline on a second benchmark; and published systems provide
cross-protocol context. Unless stated otherwise, end-to-end answer-accuracy
experiments use the same answer-generation and judging pipeline.

\subsection{Experimental Protocol}
\label{app:experimental-protocol}

\paragraph{Datasets.}
We evaluate \method{} on LoCoMo and LongMemEval-S. For LoCoMo, we report
results on both the 1,540 non-adversarial questions and the full set of 1,986
questions. For LongMemEval-S, the main experiment uses all 500 questions.

\paragraph{Answer generation and evaluation.}
For each question, the memory system first constructs the evidence supplied to
the answer model. The answer model then generates a response using the
question and the constructed evidence. For our runs, we use GPT-4o-mini with
temperature 0 and a maximum of 512 generated tokens as both the answer model
and the LLM-as-a-Judge evaluator. The same generation and judging pipeline is
used when comparing \method{} with matched evidence-construction variants.

\paragraph{Baseline comparisons.}
The external reference rows are transcribed from MemPro's GPT-4o-mini
comparison table~\citep{liu2026memproagenticmemorysystems}. MemPro reports that
this table combines baselines reproduced under its held-out protocol with
values taken from the corresponding papers. Its LoCoMo protocol reserves 154
of 1,540 questions for framework evolution and evaluates on the remaining
1,386 question-disjoint examples; its LongMemEval protocol analogously uses 50
of 500 questions for evolution and 450 for held-out evaluation. In contrast,
our \method{} rows report all 1,540 LoCoMo and all 500 LongMemEval-S questions.
The external rows are therefore contextual rather than matched comparisons.
Controlled comparisons in this paper use the same memory, question set,
answer model, and judging pipeline across evidence-construction variants.

\paragraph{Development and diagnostic subset.}
We use a fixed 100-question development and diagnostic subset to freeze the
evidence-construction configuration and conduct controlled mechanism
analyses. The subset is sampled from the full 1,986-question LoCoMo benchmark
with stratified quotas of 14 multi-hop, 16 temporal, 5 open-domain, 42
single-hop, and 23 adversarial questions. Within each category, questions are
sorted by question identifier and sampled with random seed 42, followed by a
final shuffle.

The development subset is used to fix the evidence-construction
hyperparameters before the benchmark-wide evaluations. Controlled comparisons
on this subset are intended to isolate evidence volume, temporal
reconstruction, and set-aware anchor allocation; they are not independent
benchmark estimates. The reported 1,540- and 1,986-question LoCoMo results
overlap with this development subset and should be read as complete-set
reporting, not as disjoint held-out evaluation. The 1,986-question set contains
all 100 diagnostic questions; the 1,540-question non-adversarial set contains
its non-adversarial portion.

\begin{table}[t]
\centering
\small
\renewcommand{\arraystretch}{1.05}
\begin{tabular*}{0.92\linewidth}{@{\extracolsep{\fill}}lrl@{}}
\toprule
\textbf{Evaluation set} & $\boldsymbol{N}$ & \textbf{Role in the paper} \\
\midrule
LoCoMo diagnostic & 100 & Configuration and mechanism analyses \\
LoCoMo non-adversarial & 1,540 & Aggregate benchmark result \\
LoCoMo all questions & 1,986 & Matched allocator and aggregate result \\
LongMemEval-S & 500 & One fixed user--assistant indexing configuration \\
\bottomrule
\end{tabular*}
\caption{Evaluation scope. The LoCoMo benchmark reports overlap with the
development diagnostic subset; they are not disjoint held-out estimates.}
\label{tab:evaluation-scope}
\end{table}

\FloatBarrier

\subsection{Evidence Selection Configuration}
\label{app:evidence-selection}

\paragraph{Candidate retrieval.}
At query time, \method{} first retrieves a broad candidate pool of 200
conversation turns. This pool defines the turns available to the allocator but
does not determine the final evidence view. Anchor allocation is handled by the
subsequent selection stage.

\paragraph{Anchor allocation.}
From the candidate pool, \method{} greedily selects 60 anchor turns. At each
step, the next anchor is selected according to

\begin{equation}
m_{t+1}^* =
\arg\max_{m \in \mathcal{C}_q \setminus S_t}
\left[
R(m,q)
+\lambda D(m,S_t)
+\beta C(m,q,S_t)
\right],
\label{eq:anchor-allocation-app}
\end{equation}

where $R(m,q)=s_{\mathrm{dense}}(m,q)$ is the dense retrieval score,
$D$ measures diversity with respect to the already selected set, and $C$
measures the proportion of currently uncovered query terms that occur in the
candidate. The full selector uses $\lambda=0.06$ and $\beta=0.10$.

\paragraph{Diversity.}
For a candidate $m$ and selected set $S_t$, diversity is implemented as

\begin{equation}
D(m,S_t)
=
1-
\max_{s\in S_t}
\operatorname{Jaccard}
\bigl(
\operatorname{tok}(m),
\operatorname{tok}(s)
\bigr),
\label{eq:diversity-app}
\end{equation}

with $D(m,\emptyset)=1$. The selector tokenizes each turn by lowercasing and
extracting tokens using the regular expression
\texttt{[a-z0-9']+}; $\operatorname{tok}(m)$ denotes the resulting token set.
A candidate therefore receives a lower diversity score when it substantially
overlaps with an already selected turn.

\paragraph{Query coverage.}
Query terms are formed from extracted uppercase entities, excluding stop
names, together with content words of length at least four. Candidate
matching is performed by case-insensitive substring matching. For a given
selection state, $C(m,q,S_t)$ measures the proportion of query terms not
already covered by $S_t$ that are matched by candidate $m$. If no uncovered
query terms remain, the coverage contribution is zero.

\paragraph{Selector ablations.}
The clean selector ablations activate the components as follows:

\begin{itemize}
    \item Relevance-only: $\lambda=0,\ \beta=0$;
    \item $R+D$: $\lambda=0.06,\ \beta=0$;
    \item $R+C$: $\lambda=0,\ \beta=0.10$;
    \item $R+D+C$: $\lambda=0.06,\ \beta=0.10$.
\end{itemize}

All selector variants use the same candidate pool, anchor budget, and
temporal reconstruction procedure unless otherwise specified.

\paragraph{Ordering and ties.}
The greedy allocator uses a strict greater-than comparison, so an exact score
tie retains whichever tied candidate is encountered first.

\begin{table}[t]
\centering
\small
\renewcommand{\arraystretch}{1.05}
\begin{tabular*}{0.88\linewidth}{@{\extracolsep{\fill}}ll@{}}
\toprule
\textbf{Setting} & \textbf{Value} \\
\midrule
Dense retrieval & \texttt{text-embedding-3-small}; Qdrant \\
Candidate pool / anchor budget & 200 / 60 turns \\
Allocator weights $(\lambda,\beta)$ & $(0.06, 0.10)$ \\
Same-session neighborhood & Two turns per side \\
Answer model / judge & GPT-4o-mini / GPT-4o-mini \\
Temperature / maximum output & 0 / 512 tokens \\
\bottomrule
\end{tabular*}
\caption{Configuration used for the main \method{} evaluations.}
\label{tab:main-configuration}
\end{table}

\FloatBarrier

\subsection{Same-Session Source-Context Reconstruction}
\label{app:temporal-reconstruction}

The selected anchors are not directly passed to the answer model. For each
anchor $m_i$, \method{} retrieves the $w$ preceding and following turns from
the same conversation session:

\begin{equation}
\mathcal{N}_w(m_i)
=
\{m_j \mid s_j=s_i,\ |\tau_j-\tau_i|\leq w\},
\label{eq:temporal-neighborhood-app}
\end{equation}

where $s_i$ denotes the session identifier and $\tau_i$ denotes the
session-local turn position. We use $w=2$ in all main experiments.

Reconstruction operates on the original stored conversational
memory rather than performing another retrieval over the candidate pool.
Consequently, a same-session neighbor does not need to appear among the initial
candidate turns. The retrieved neighborhoods are merged and deduplicated,
and the retained turns are restored to their original chronological order
before being passed to the answer model.

This distinction separates anchor allocation from context reconstruction:
the anchor set determines which locations in the conversation are considered
important, while reconstruction restores nearby dialogue needed to
interpret those locations.

\subsection{Evidence-Volume Controls}
\label{sec:evidence-volume}

We use several controls to distinguish the effect of temporal reconstruction
from simply providing more retrieved turns.

\paragraph{Top-300 Raw.}
Top-300 Raw supplies the 300 highest-ranked conversation turns directly,
without temporal expansion.

\paragraph{Random Expansion.}
Random Expansion starts from the same relevance-only anchors used by the
temporal reconstruction condition and adds randomly sampled non-anchor turns
from the full conversation until the final number of entries approximately
matches the reconstructed context. The resulting context contains
approximately 203 entries on average. This control matches evidence-entry
count rather than exact token count.

\paragraph{Token-matched random expansion.}
This control starts from the same relevance-only anchors as temporal
reconstruction and does not change them. It repeatedly samples non-anchor turns
from the full conversation until the resulting evidence approximately matches
the token volume of the temporal-reconstruction condition. We run five
independent seeds (42--46) on the 100-question development diagnostic and
report their mean accuracy, 68.4\%. This comparison tests whether local
same-session structure contributes beyond the amount of evidence supplied to
the answer model.

\subsection{LongMemEval-S Protocol}
\label{app:longmemeval-protocol}

The main LongMemEval-S result comes from one fixed configuration applied to all
500 questions, with both user and assistant turns indexed. The answer model,
prompt, and evaluation procedure are also fixed across question types. This
run achieves 86.6\% accuracy (433/500). A separate 56-question role-coverage
diagnostic compares user-only indexing with user--assistant indexing; its
results appear in Appendix~\ref{app:longmemeval-breakdown} and are not used to
construct the 500-question result.

\FloatBarrier
\subsection{Evidence Budget and Context Volume}
\label{sec:evidence-budget}

To examine whether the improvement of \method{} is explained simply by
providing more retrieved evidence, we compare four evidence-construction
conditions on the fixed 100-question diagnostic subset of LoCoMo.
Table~\ref{tab:evidence_budget} in the main paper reports the accuracy and
context volume for these conditions. The four conditions provide three
controlled contrasts. Dense-60 versus
Dense-200 measures broader independent retrieval (+6 points). Dense-200 versus
Dense + Temporal compares different evidence organization at similar final
volume (+16 points). Dense + Temporal versus \method{} changes only anchor
allocation at nearly unchanged context volume (+4 points). These development
diagnostics are not an additive decomposition of benchmark-wide performance.

The matched evidence volume in this experiment refers to the retrieved
conversation content rather than the complete API prompt. The latter also
contains evaluation instructions and other prompt tokens and is therefore
larger than the retrieved evidence itself.

\begin{table}[t]
\centering
\small
\renewcommand{\arraystretch}{1.05}
\begin{tabular*}{0.94\linewidth}{@{\extracolsep{\fill}}>{\raggedright\arraybackslash}p{0.30\linewidth}>{\raggedright\arraybackslash}p{0.27\linewidth}>{\raggedright\arraybackslash}p{0.29\linewidth}@{}}
\toprule
\textbf{Comparison} & \textbf{Held fixed or matched} & \textbf{Question isolated} \\
\midrule
Relevance-only vs. \method{} & Memory, questions, candidate pool, anchor
budget, reconstruction, model, judge & Does set-aware allocation improve answers? \\
Dense + Temporal vs. \method{} & Candidate pool, anchor budget, reconstruction
procedure, approximate context volume & Does allocation help beyond local reconstruction? \\
Random vs. temporal expansion & Same relevance anchors; approximate turn count
& Does local context help beyond more turns? \\
Token-matched random vs. temporal & Same relevance anchors; approximate
evidence-token volume & Does local context help beyond more text? \\
\bottomrule
\end{tabular*}
\caption{Purpose of the controlled comparisons. Approximate matching is stated
explicitly; only the allocator comparison over all 1,986 questions is the
benchmark-wide matched test.}
\label{tab:control-map}
\end{table}

\subsection{Query-Time Efficiency}
\label{sec:efficiency-details}

We measure query-time memory-construction latency on the same conversational
memory under 100 diagnostic queries. The optimized implementation precomputes
the token sets and pairwise Jaccard similarities used by the selector. The
selection comparison and tie-breaking rule remain unchanged.

\begin{table}[t]
\centering
\small
\renewcommand{\arraystretch}{1.05}
\begin{tabular}{lrrrr}
\toprule
\textbf{Method}
& \textbf{Retrieval}
& \textbf{Selection}
& \textbf{Reconstruction}
& \textbf{Total} \\
\midrule
Dense-60
& 21
& --
& --
& 21 \\
\method{} (original)
& 30
& 7046
& 108
& 7183 \\
\method{} (optimized)
& 25
& 115
& 122
& 262 \\
\bottomrule
\end{tabular}
\caption{Mean query-time memory-construction latency in milliseconds per query
over 100 diagnostic questions. Stage means are rounded independently; Total
is measured end to end. Caching changes the implementation but not the scoring
rule.}
\label{tab:latency}
\end{table}

The original implementation spends most of its time repeatedly tokenizing
turns and computing Jaccard overlap during greedy selection. Caching these
representations reduces selection latency from 7.05\,s to 115\,ms per query,
reducing total memory-construction latency from 7.18\,s to 262\,ms. Relative
to Dense-60, the optimized implementation therefore introduces
approximately 241\,ms of additional memory-organization overhead per query.

The optimization changes only the implementation of the selector rather than
its scoring rule or tie-breaking rule. The aggregate evidence statistics over
the 100 diagnostic queries remain unchanged at 202.82 turns and 4,574.87
words.

\subsection{API Context and Token Usage}
\label{sec:api-usage}

We additionally record representative API usage statistics for the dense
baseline and \method{} on the diagnostic queries.

\begin{table}[t]
\centering
\small
\renewcommand{\arraystretch}{1.05}
\begin{tabular}{lrr}
\toprule
& \textbf{Dense-60} & \textbf{\method{}} \\
\midrule
Retrieved turns
& 60.0
& 202.82 \\
Retrieved words
& 1{,}268
& 4{,}575 \\
Prompt tokens
& 4{,}206
& 13{,}218 \\
Completion tokens
& 129.4
& 111.1 \\
Total tokens
& 4{,}335
& 13{,}330 \\
\bottomrule
\end{tabular}
\caption{Per-query API usage statistics over the 100 diagnostic questions.
Prompt-token counts include the surrounding evaluation prompt and are
therefore not equivalent to retrieved-evidence tokens.}
\label{tab:api-usage}
\end{table}

The larger input prompt of \method{} primarily reflects the larger final
evidence context. Dense retrieval supplies approximately 60 turns and 1,268
words, whereas temporal reconstruction in \method{} produces approximately
203 turns and 4,575 words. The completion length is similar across the two
conditions, with 129.4 tokens for Dense and 111.1 tokens for \method{}.

\subsection{Paired Uncertainty and Significance}
\label{sec:significance}

We compare \method{} with the matched relevance-only allocator on the same
1,986 LoCoMo questions. \method{} corrects 135 questions that the baseline
answers incorrectly and loses 98 questions that the baseline answers
correctly, giving a net gain of

\begin{equation}
100\times\frac{135-98}{1986}=1.86
\end{equation}

percentage points.

\paragraph{Primary cluster uncertainty.}
Because multiple questions belong to the same conversations, the primary
uncertainty estimate is a conversation-cluster bootstrap over the 10 LoCoMo
conversations. Each of 10,000 replicates samples conversations with replacement,
retains their paired question outcomes, and recomputes the accuracy difference.
The 95\% percentile interval, using the 2.5th and 97.5th percentiles, is
[0.59, 3.15] percentage points. Pairing is preserved because both systems
answer the same questions under the same generation and judging pipeline.

\paragraph{Question-level paired checks.}
The continuity-corrected McNemar statistic is

\begin{equation}
\frac{(|135-98|-1)^2}{135+98}=5.562,
\end{equation}

corresponding to a two-sided $p=.018$. For reference, a paired question-level
bootstrap gives a 95\% interval of [0.40, 3.37] percentage points. These
question-level checks do not replace the conversation-cluster interval.

\FloatBarrier
\section{Additional Results}
\label{sec:additional-results}
\setcounter{topnumber}{3}

\subsection{LongMemEval-S Role Coverage}
\label{app:longmemeval-breakdown}

\begin{table}[htbp]
\centering
\scriptsize
\renewcommand{\arraystretch}{1.04}
\setlength{\tabcolsep}{4pt}
\begin{tabular}{lrr}
\toprule
\textbf{Indexing} & $N$ & \textbf{Accuracy} \\
\midrule
User-only & 56 & 19.6 \\
\rowcolor{green!8}
Dual-role & 56 & \textbf{78.6} \\
\bottomrule
\end{tabular}
\caption{Role coverage on 56 assistant-memory questions (\%).}
\label{tab:longmemeval-role}
\end{table}

The main LongMemEval-S result uses a single dual-role indexing configuration
applied uniformly to all 500 questions, where both user and assistant turns
are searchable. This configuration achieves 86.6\% accuracy (433/500).

We additionally examine the role of assistant-side memory using the 56
single-session assistant questions. Under user-only indexing, the system
answers 11/56 questions correctly (19.6\%). When both user and assistant
turns are searchable, the same 56-question subset reaches 44/56 (78.6\%).

The 58.9 percentage-point difference indicates that restricting retrieval to
user turns can prevent access to information explicitly provided by the
assistant. This diagnostic is not used to construct the overall 86.6\%
result; the main result is obtained from the single dual-role indexing
configuration applied uniformly to all 500 questions.

\subsection{Temporal Reconstruction Controls}
\label{sec:temporal-controls}

On the fixed 100-question diagnostic set, the evidence-volume controls give
the following results:

\begin{table}[t]
\centering
\small
\renewcommand{\arraystretch}{1.05}
\begin{tabular*}{0.82\linewidth}{@{\extracolsep{\fill}}lrr@{}}
\toprule
\textbf{Condition} & \textbf{Accuracy} & \textbf{$\Delta$} \\
\midrule
\multicolumn{3}{l}{\textit{Reconstruction and volume controls}} \\
Top-300 Raw & 63.0 & -- \\
Random Expansion & 67.0 & -- \\
Token-matched Random Expansion & 68.4 & -- \\
Temporal Reconstruction & \textbf{73.0} & -- \\
\addlinespace[2pt]
\multicolumn{3}{l}{\textit{Allocator ablations}} \\
Relevance-only & 73.0 & -- \\
$R+D$ & 70.0 & -3.0 \\
$R+C$ & 73.0 & 0.0 \\
\rowcolor{green!8}
$R+D+C$ (\method{}) & \textbf{77.0} & +4.0 \\
\bottomrule
\end{tabular*}
\caption{Reconstruction controls and allocator ablations on the fixed
100-question diagnostic set (\%). Token-matched Random Expansion reports the
mean over seeds 42--46. All allocator variants share candidates, budget, and
reconstruction.}
\label{tab:temporal-controls}
\end{table}

The temporal condition achieves higher accuracy than raw top-ranked retrieval
and both random-expansion controls. In particular, approximately matching the
token volume with random expansion does not reproduce the temporal result.
\subsection{Selector Ablations}
\label{sec:selector-ablations}

We evaluate the components of the anchor allocator on the fixed 100-question
diagnostic set. All variants retain relevance and differ only in whether they
include the diversity term $D$, the coverage term $C$, or both.

Among the tested variants, $R+D$ and $R+C$ do not improve answer accuracy over
relevance-only allocation, while $R+D+C$ improves it from 73.0\% to 77.0\%.
Thus, only the combined rule reproduces the diagnostic gain; the complete-set
matched experiment carries the benchmark-wide allocator claim.

\FloatBarrier
\subsection{Pre-Reconstruction Anchor Availability}
\label{sec:evidence-availability}

\begin{table}[htbp]
\centering
\scriptsize
\renewcommand{\arraystretch}{1.04}
\setlength{\tabcolsep}{3pt}
\begin{tabular}{lrr}
\toprule
\textbf{Outcome} & \textbf{Count} & \textbf{Fraction} \\
\midrule
Gold items & 142 & 100.0 \\
Selected anchor & 84 & 59.2 \\
Candidate, not anchor & 41 & 28.9 \\
Absent from candidates & 17 & 12.0 \\
\bottomrule
\end{tabular}
\caption{Gold-evidence status before reconstruction (\%).}
\label{tab:evidence-availability}
\end{table}

We characterize gold-evidence status at the anchor stage, before temporal
reconstruction. On the diagnostic set, 142 dataset-provided gold evidence
items are identified. The initial candidate pool contains
125 of these items, while 17 gold items are absent from the candidate pool.
Of the gold items available in the candidate pool, 84 are selected as anchors
and 41 are not.

Among the 58 gold items not selected as anchors, 41 (70.7\%) are already
present in the initial candidate pool and visible to the allocator. Candidate
access therefore leaves an allocation opportunity on this diagnostic, although
17 gold items remain unavailable. Because reconstruction reads the original memory,
an item not selected as an anchor may still enter the final context as a
neighbor; the table therefore describes the anchor stage only.

\subsection{Additional Allocation Comparisons}
\label{sec:additional-allocation}

\begin{table}[htbp]
\centering
\scriptsize
\renewcommand{\arraystretch}{1.04}
\setlength{\tabcolsep}{4pt}
\begin{tabular}{lr}
\toprule
\textbf{Allocation rule} & \textbf{Accuracy} \\
\midrule
Relevance-only & 73.0 \\
Protected coverage & 72.0 \\
Protected $D+C$ & 73.0 \\
\rowcolor{green!8}
\method{} (unprotected $R+D+C$) & \textbf{77.0} \\
Gold-informed reference & 78.0 \\
\bottomrule
\end{tabular}
\caption{Allocation controls on 100 diagnostic questions (\%). The
gold-informed reference is not deployable.}
\label{tab:additional-allocation}
\end{table}

We compare additional allocation rules on the same fixed 100-question
diagnostic set.

The protected variants retain a prefix of the dense ranking while applying
additional selection criteria to the remaining budget. In this diagnostic,
protecting the prefix does not improve over relevance-only selection. The
gold-informed allocator provides a reference point for the remaining
allocation headroom, but uses gold evidence identities and is therefore not
deployable.

\subsection{Selection-Only Iterative Evidence Acquisition}
\label{sec:agentic-extension}

We additionally evaluate a gap-directed acquisition variant only as a
selection-level diagnostic; it is not part of any end-to-end \method{} result.
Given an initial evidence state, the procedure
iteratively identifies unresolved evidence gaps, scores candidate turns with
respect to the current gap, and updates the evidence state. The candidate
pool itself is fixed throughout the procedure; thus, this extension does not
introduce an additional retrieval stage or expand the set of retrievable
turns.

The temporary evidence state records selected evidence, grounded entities
and events, temporal matches, unresolved information gaps, and unsatisfied
query terms. We consider several types of unresolved gaps, including missing
entities, missing facts or events, missing temporal relations, missing
multi-hop bridges, and insufficient evidence for multi-faceted questions.
Given an unresolved gap $g$, a candidate $m$ is scored by extending the base
allocation rule with gap-conditioned relevance and temporal usefulness:

\begin{equation}
u_t^{\mathrm{gap}}(m;q,g)
=
R(m,q)
+\lambda D(m,S_t)
+\beta C(m,q,S_t)
+\gamma G(m,g)
+\delta T(m,S_t),
\label{eq:iterative-score-app}
\end{equation}

where $G(m,g)$ measures the relevance of candidate $m$ to the unresolved gap,
and $T(m,S_t)$ measures its temporal usefulness for the current evidence state.
For this diagnostic, we use $\lambda=0.06$, $\beta=0.12$,
$\gamma=0.18$, and $\delta=0.08$.

The procedure initializes the evidence state with two anchors and performs at
most 80 acquisition steps, subject to a maximum of 60 anchors. For
multi-faceted questions, the additional-evidence requirement is
$\max(3,\min(8,2+|\mathcal{T}_q|))$, where $\mathcal{T}_q$ denotes the query
facets tracked in the temporary evidence state. At each step, the procedure selects
evidence for an unresolved gap and updates the evidence state. If the
procedure terminates before the 60-anchor budget is exhausted, the remaining
slots are filled from the candidate pool so that the final anchor budget is
comparable to the fixed-budget base selector. The gap-conditioned terms are
used only in this iterative diagnostic; the main \method{} configuration
does not require an iterative acquisition loop.

\paragraph{Evidence-coverage metrics.}
Let $G_q$ denote the dataset-provided gold evidence items for question $q$,
and let $S_q^{60}$ denote the final set of at most 60 selected anchors. We
measure Gold Recall@60 as the fraction of gold evidence items directly
selected as anchors:

\begin{equation}
\mathrm{GoldRecall@60}
=
\frac{1}{N}\sum_{q=1}^{N}
\frac{|G_q\cap S_q^{60}|}{|G_q|}.
\label{eq:gold-recall}
\end{equation}

We additionally measure Full-Gold Coverage after temporal reconstruction.
Gold Recall@60 is the mean per-question item recall, whereas Full-Gold Coverage
is a question-level all-evidence criterion: it asks whether all
dataset-provided gold evidence for a question is contained in the final
reconstructed evidence context $\mathcal{E}_q$:

\begin{equation}
\mathrm{FullGoldCoverage}
=
\frac{1}{N}
\sum_{q=1}^{N}
\mathbf{1}[G_q\subseteq\mathcal{E}_q].
\label{eq:full-gold}
\end{equation}

Thus, Gold Recall@60 averages direct anchor recall across questions, whereas
Full-Gold Coverage measures the fraction of questions whose complete gold
evidence set is recovered after temporal reconstruction. The pooled 84/142
anchor count in Table~\ref{tab:evidence-availability} uses a different
aggregation and is not directly comparable to mean Gold Recall@60.

On the fixed 100-question diagnostic set, iterative acquisition increases
Gold Recall@60 from 85.2\% to 86.6\% and Full-Gold Coverage from 84.0\% to
87.0\% relative to the fixed-budget base selector. We additionally report an
offline gold-informed planner as a reference.
This planner directly uses gold evidence identities to
guide selection within the candidate pool and is therefore not deployable.
It reaches 90.8\% Gold Recall@60 and 90.0\% Full-Gold Coverage.

These results are selection-level diagnostics rather than end-to-end
generative evaluation. They indicate that unresolved evidence gaps can
provide useful signals for reallocating a fixed anchor budget, but they do
not establish an answer-accuracy improvement from adding an iterative agent
loop to the complete memory system. We therefore treat iterative evidence
acquisition as an exploratory extension rather than a component of the main
\method{} evaluation.

\end{document}